# What can Computer Vision learn from Ranganathan?

Mayukh Bagchi[1][0000-0002-2946-5018] and Fausto Giunchiglia[1][0000-0002-5903-6150]

[1]DISI, University of Trento, Italy.
{mayukh.bagchi,fausto.giunchiglia}@unitn.it

**Abstract** The Semantic Gap Problem (SGP) in Computer Vision (CV) arises from the misalignment between visual and lexical semantics leading to flawed CV dataset design and CV benchmarks. This paper proposes that classification principles of S.R. Ranganathan can offer a principled starting point to address SGP and design high-quality CV datasets. We elucidate how these principles, suitably adapted, underpin the *vTelos* CV annotation methodology. The paper also briefly presents experimental evidence showing improvements in CV annotation and accuracy, thereby, validating *vTelos*.

## 1 Introduction

The SGP [1], the unavoidable misalignment between visual information encoded in images and linguistic labels used to describe them, is an established bottleneck in designing CV benchmark datasets. We argue that the root cause of the SGP is an *unprincipled annotation process* that allows arbitrary subjective classification decisions while designing such benchmarks and results in CV datasets (e.g., ImageNet [2]) with systematic flaws [3]. The arbitrariness stems from an unspecified *many-to-many mapping* between images and labels [4-6], where an image can justify multiple labels and a label can describe visually diverse images (see, Figure 1, for examples of different categories of images including single-object, multi-object and mislabelled images, reproduced from [7]). These flaws, in turn, are critical as they can undermine the reliability and generalizability of multimodal Artificial Intelligence (AI) systems, e.g., Vision Language Model (VLMs), trained and evaluated on such data.

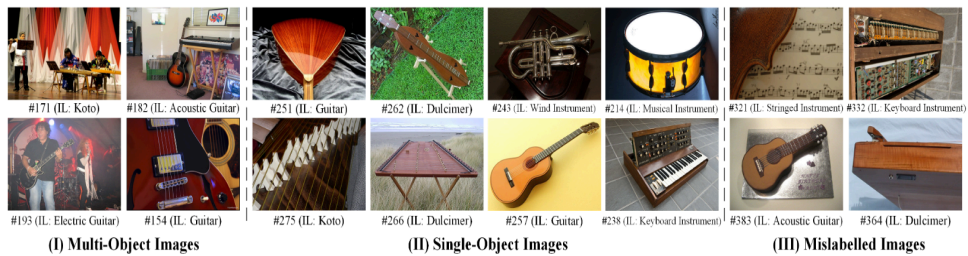

**Figure 1**: Samples from ImageNet Musical Instrument Sub-Hierarchy (IL stands for *ImageNet Label*).

We posit a shift from *ad-hoc labeling-based* CV annotation to *principled classification-based* CV annotation. To that end, this paper advances the proposal that the CV dataset design can find a tried-and-tested starting point in the principles of library classification as advanced by S.R. Ranganathan [8,9]. The analytico-synthetic classification paradigm, originally developed for organizing knowledge in libraries, provides a stratified process that characteristically decouples perception, conception, lexicalization and notation, while, at the same time, keeps them functionally interlinked [10,11]. This process offers a base for aligning visual semantics (derived from visual properties) with lexical semantics (derived from word meanings) [12,13], thereby, absolving the SGP.

In the above context, we demonstrate how Ranganathan's principles can be adapted into a principled classification-based CV annotation methodology termed *vTelos* [7]. To that end, *vTelos* enforces a *one-to-one alignment* between images and labels by grounding the annotation process in visual properties and lexical-semantic definitions. The remainder of the paper is organized as follows. Sections (2) and (3) elucidates this cross-disciplinary adaptation in terms of library classification and visual classification, respectively. Section (4) presents experimental evidence that *vTelos* significantly improves CV annotation and accuracy. Section (5) concludes the paper.

## 2 Ranganathan's Theory of Classification

The theory of classification advanced by S.R. Ranganathan offers a tried-and-tested framework for systematic knowledge organization [14-16]. To that end, the analytico-synthetic paradigm of classification was developed to manage the complexity of multidimensional (bibliographic) knowledge by isolating the process of classification into distinct yet functionally linked stages. This structured approach, in addition to the body of classificatory guiding principles (*'canons'*), provided a powerful model to transform bibliographic thought content into well-defined classification categories.

A key characteristic of this paradigm is the stratification of classification work across four planes, namely, the *Pre-Idea Stage*, the Idea Plane, the Verbal Plane and the Notational Plane. The Pre-Idea stage concerns the initial perception and recognition of entities. It is where pure percepts aggregate into compound percepts to finally form basic concepts. In CV terms, this stage aligns with the extraction of visual properties and the formation of substance concepts from images (or videos). Subsequently, the Idea Plane concerns organizing these recognized concepts into a classification hierarchy based on their (inherent) characteristics. This is achieved via the cumulative application of canons of classification of the Idea Plane, wherein, concepts, at each level, are defined by a general category (*genus*) and specific distinguishing features (*differentia*). For CV data, this stage translates to building a visual classification hierarchy using visual properties.

The final two planes of work bridge perception with language and notation. To that end, the Verbal Plane assigns standardized linguistic labels to the nodes of the classification hierarchy (visual hierarchy, in the case of CV), thereby, creating a lexical (subsumption) hierarchy. Finally, the Notational Plane disambiguates these linguistically labeled concepts by attaching unique identifiers (e.g., classification number) resolving issues of synonymy and polysemy. The entire classification process across the planes of work is governed by laws and canons. In particular, the *Law of Local Variation* ensures that a classification is adapted for a specific context and purpose. This principle is key to CV's need for context-aware and egocentric object recognition benchmarks.

## 3 Visual Classification

In visual classification the Pre-Idea Stage of Ranganathan corresponds to the *Substance Concept Recognition* phase. In this phase, a CV system, much like a human observer, *encounters* visual data from a cumulation of *partial views* (e.g., of a musical instrument). To that end, it extracts low-level visual properties (pure percepts, as from Ranganathan) which, over multiple encounters, aggregate into compound percepts, thereby forming (stable) substance concepts, e.g., photos of *stringed instrument*, *stringed instrument with six taut strings* (see, Figure 1). The process is computationally supported by a cumulative memory, namely, a dynamic knowledge base [17] that incrementally refines concepts with new visual evidence, thereby, providing a mechanism for persistent and evolving CV object identity.

The output of this recognition phase feeds is the input to the next phase of *Visual Classification*. In this phase, the (isolated) substance concepts are collected and organized into a purpose-specific visual subsumption hierarchy. This is implemented by applying the definition of (visual) *genus-differentiae* to the identified visual properties. The *genus* represents a shared visual characteristic that groups visual concepts (e.g., *has taut strings*), while the *differentia* provides the specific visual property that distinguishes them within that group (e.g., *has six taut strings* versus *has thirteen taut strings*). By iteratively applying this logic, this phase archives a visual classification hierarchy. For example, a root visual concept like *musical instrument* is classified into *stringed instrument* and *wind instrument* based on the *genus* of sound production, and *stringed instrument* is further classified into *guitar* and *koto* based on the differentia of string counts. This hierarchical construction is based on visual affordances and it ensures the organization is grounded in visual perception rather than linguistic labels.

The transition from a visual hierarchy to a lexical-semantic hierarchy is implemented in the next phase of *Linguistic Classification*. The nodes of the visual subsumption hierarchy, each representing a substance concept defined by visual properties, is labeled for human and machine reasoning. To that end, it involves mapping each node to a word or synset from a natural or domain language (often, exploiting multilingual lexical-semantic resources [18,19]). For instance, the substance concept defined by the visual differentia *has six strings* is assigned the label *guitar*. This step creates a lexical subsumption hierarchy out of the visual classification hierarchy. Notice that this process must account for: (i) lexical gaps, wherein, a language may lack a direct label for a visual concept, and (ii) egocentric purpose, where the chosen label may reflect a specific user's context. The alignment, therefore, is definitional, i.e., the linguistic label's gloss must linguistically encode the visual genus and differentia that defined the node, ensuring a one-to-one mapping between the visual and lexical concept at that hierarchical level.

The final phase of *Conceptual Classification* introduces a layer of formal disambiguation essential for computational reasoning in the context of CV-driven applications. This is important as language is inherently ambiguous due to linguistic phenomena like synonymy and polysemy. To eliminate this ambiguity, each linguistically labeled concept in the lexical hierarchy is assigned a unique supralingual identifier (again, exploiting multilingual lexical-semantic resources). For example, the classification concept labeled *guitar* might be assigned an identifier like *1278956*. This identifier is independent of any particular language, thereby ensuring that the concept remains distinct whether it is referred to as *guitar* or *six-stringed instrument*. This phase finalizes the transformation of a perceptual *substance concept* into a formal, machine-actionable *classification concept*, ready for use in CV-driven knowledge graphs, databases and reasoning systems.

To operationalize this adapted process as a real-world annotation pipeline, we developed the *vTelos* methodology and a redefinition of CV annotator roles. *vTelos* structures the CV annotation task around four key design choices: C1) Object Localization, C2) Visual Classification, C3) Label Generation, and C4) Label Disambiguation. Notice that differently from mainstream CV annotation practice, in *vTelos*, the linguistic choices (C3 and C4) are made prior to the visual ones (C1 and C2). This means the space of possible labels, each with a clearly defined meaning from a multilingual lexical-semantic resource, is established which creates a controlled vocabulary. The annotator then selects labels based on how well definitions match the visual properties of the localized object.

This workflow necessitates splitting the traditional *annotator* role into two specialized functions borrowed from knowledge organization: the *Classificationist* and the *Classifier*. The classificationist is the domain expert who performs *a priori* work of linguistic choices C3 and C4. They define the lexical-semantic hierarchy and ensure each label's gloss provides clear linguistic genus-differentia.. The classifier, who could be a human annotator or a (semi-)automated system [20], then performs the perceptual work of visual choices C1 and C2. To that end, they localize objects in images and, by matching visual properties to the pre-defined linguistic differentia, assign the appropriate label from the classificationist's schema. This separation of linguistic and visual choices ensure that subjective interpretation is minimized during annotation. The classifier does not choose a label based on intuition, but executes a match between visual evidence and a pre-established semantic definition. The *vTelos* methodology, therefore, enforces a principled, one-to-one mapping between visual and lexical semantics, directly tackling the many-to-many mappings at the heart of the SGP.

## 4 Experimental Evidence

We applied the *vTelos* methodology to a subset - the musical instrument hierarchy - of ImageNet. The experiment [7] was designed to compare mainstream ad-hoc label-based annotation against the principled classification-based annotation. To that end, the subset was used to design a ground truth dataset and annotators were divided into groups: (i) one group annotated images via ad-hoc selection from ImageNet category labels, while (ii) the second group performed annotation using the *vTelos* methodology before assigning a final label. The results demonstrated an 18% improvement in inter-annotator agreement for (ii) over (i) [7]. This indicates a reduction in subjective judgment as annotators converge more reliably with respect to visual properties than labels. Further, when the full *vTelos* process including object localization was implemented, agreement amongst expert annotators reached near-perfect levels. This confirms that structuring the CV annotation task with principled classification choices absolves SGP ambiguity.

The impact on CV accuracy was equally encouraging. Several state-of-the-art convolutional neural networks, including ResNet and VGG, were trained on the dataset *re-annotated* following the *vTelos* methodology. In comparison to models trained on the original ImageNet labels, these models showed improvement in CV classification accuracy, with gains of up to 23% for some architectures [7]. Additionally, an ablation study confirmed that the key driver of this improvement was the *Visual Classification* phase itself, rather than merely the isolation of single objects. The models effectively learned to associate the linguistically grounded differentia with the corresponding visual features, as evidenced by *attention heatmaps* that highlighted relevant image regions [7]. This case study provided concrete evidence that adopting principles from Ranganathan's library science directly addresses the SGP and produces high-quality CV datasets.

## 5 Conclusion

In summary, the CV community can learn a key lesson from Ranganathan, namely, that principled classification is foundational for high-quality CV dataset design.